\documentclass{article}
\usepackage{ijcai16}

\usepackage{times}

\usepackage{graphicx}

\usepackage{color, soul}
\usepackage{bbm}
\usepackage{amsmath,amssymb}
\usepackage{bm}
\usepackage{caption}

\DeclareMathOperator*{\argmax}{\arg\!\max}
\usepackage{hhline}
\usepackage{subcaption}
\usepackage{url}

\usepackage{algorithm}
\usepackage{algpseudocode}
\usepackage{multirow}

\usepackage{booktabs} 
\usepackage{array} 
\usepackage{paralist} 
\usepackage{verbatim} 
\usepackage{dsfont}

\title{Learning Social Affordance for Human-Robot Interaction}
\author{Tianmin Shu$^1$, M. S. Ryoo$^2$ and Song-Chun Zhu$^1$ \\ 
$^1$ Center for Vision, Cognition, Learning and Autonomy, University of California, Los Angeles \\
$^2$ School of Informatics and Computing, Indiana University, Bloomington \\
{tianmin.shu@ucla.edu\quad mryoo@indiana.edu\quad sczhu@stat.ucla.edu}\\ 
}

\def \db {\bm{d}} 

\begin{document}

\maketitle

\begin{abstract}

In this paper, we present an approach for robot learning of \emph{social affordance} from human activity videos. We consider the problem in the context of human-robot interaction: Our approach learns structural representations of human-human (and human-object-human) interactions, describing how body-parts of each agent move with respect to each other and what spatial relations they should maintain to complete each sub-event (i.e., sub-goal). This enables the robot to infer its own movement in reaction to the human body motion, allowing it to naturally replicate such interactions.

We introduce the representation of \emph{social affordance} and propose a generative model for its weakly supervised learning from human demonstration videos. Our approach discovers critical steps (i.e., latent sub-events) in an interaction and the typical motion associated with them, learning what body-parts should be involved and how. The experimental results demonstrate that our Markov Chain Monte Carlo (MCMC) based learning algorithm automatically discovers semantically meaningful social affordance from RGB-D videos, which allows us to generate appropriate full body motion for an agent.

\end{abstract}

\begin{figure}[t]
    \centering
    \includegraphics[width=1.0\linewidth]{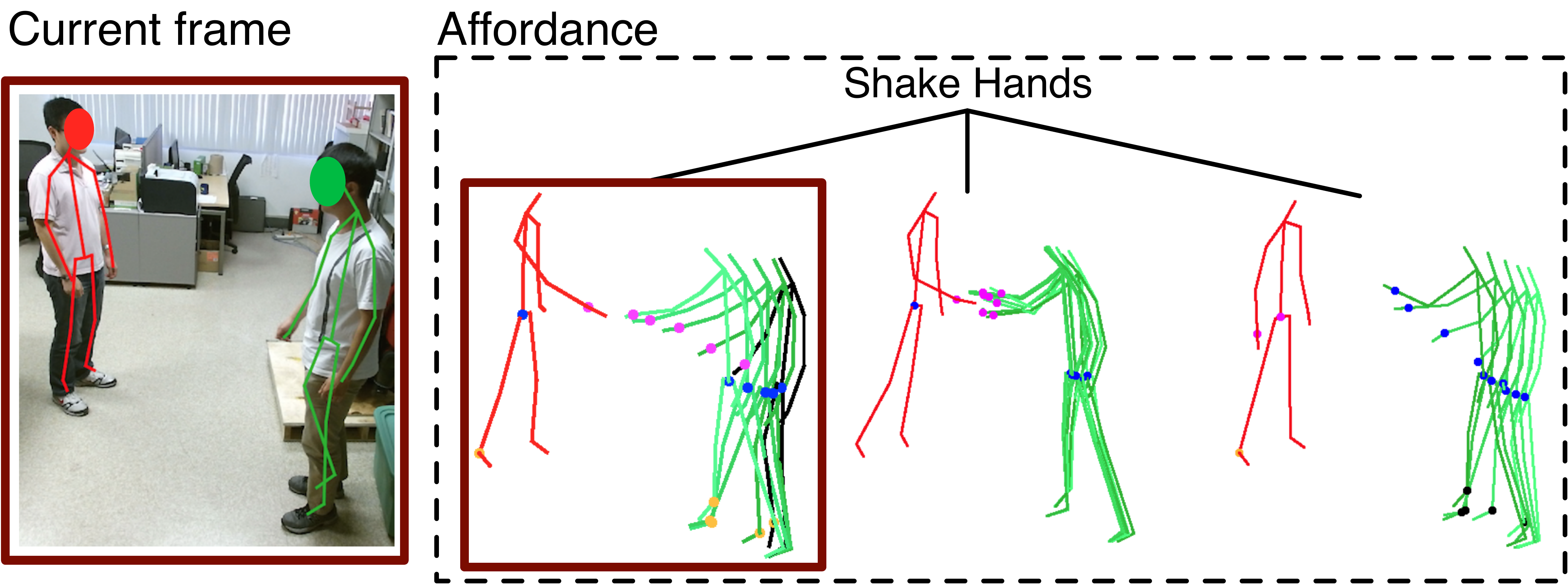}
\caption{Visualization of our social affordance. The green (right) person is considered as our agent (e.g., a robot), and we illustrate (1) what sub-event the agent needs to do given the current status and (2) how it should move in reaction to the red (left) person's body-parts to execute such sub-event. The black skeleton indicates the current frame estimation, and greens are for future estimates. The right figure shows a hierarchical activity affordance representation, where affordance of each sub-event is described as the motion of body joints. We also visualize the learned affordable joints with circles, and their grouping is denoted by the colors. Note that the grouping varies in different sub-events.}
\label{fig:intro}
\end{figure}


\section{Introduction}



The concept of ``affordance learning'' is receiving an increasing amount of attention from robotics, computer vision, and human-robot interaction researchers. The term \emph{affordance} was originally defined as ``action possibilities'' of things (e.g., objects, environments, and other agents) by \cite{Gibson}, and it has attracted researchers to study computational modeling of such concept \cite{Montesano2008,Gupta2011,Kjellstrom2011,Bogdan2012,Jiang2013,Zhu2014,Koppula2014,Pieropan2014,Pieropan2015,Sheng2015,Zhu2015}. The idea behind modern affordance learning research is to enable robot learning of ``what activities are possible'' (i.e., semantic-level affordances) and ``where/how it can execute such activities'' (i.e., spatial-level and motion-level affordances) from human examples. Such ability not only enables robot planning of possible actions, but also allows robots to replicate complicated human activities. Based on training videos of humans performing activities, the robot will infer when particular sub-events can be executed and how it should move its own body-parts in order to do so.

So far, most previous works on robot affordance learning have only focused on the scenario of a single robot (or a single human) manipulating an object (e.g., \cite{Koppula2014}). These systems assumed that affordance solely depends on the spatial location of the object, its trajectory, and the intended action of the robot. Consequently affordance was defined as a unary function in the sense that there is only one agent (i.e., the robot) involved.

However, in order for a robot to perform collaborative tasks and \emph{interact} with humans, computing single-robot object manipulation affordances based on object recognition is insufficient. In these human-robot interaction scenarios, there are multiple agents (humans and robots) in the scene and they interact and react. Thus, the robot must (1) represent its affordance as ``interactions'' between body joints of multiple agents, and (2) learn to compute such hierarchical affordances based on its status. Its affordance should become activated only when the action makes sense in the social context. For instance, the fact that human's hand is a location of affordance doesn't mean that the robot can grab it whenever it feels like. The robot should consider grabbing the human hand only when the person is interested in performing hand-shake activity with it. 



Therefore, in this paper, we introduce the new concept of \emph{social affordances}, and present an approach to learn them from human videos. We formulate the problem as the learning of structural representations of social activities describing how the agents and their body-parts move. Such representation must contain a sufficient amount of information to execute the activity (e.g., how should it be decomposed? what body-parts are important? how should the body-parts move?), allowing its social affordance at each time frame to be computed by inferring the status of the activity and by computing the most appropriate motion to make the overall activity successful (Figure~\ref{fig:intro}). Since we consider the problem particularly in the context of human-robot interaction, activity representation involving multiple agents with multiple affordable body-parts must be learned, and the inference on a robot's affordance should be made by treating it as one of the agents.


Our problem is challenging for the following reasons: (i) human skeletons estimated from RGB-D videos are noisy due to occlusion, making the learning difficult; (ii) human interactions have much more complex temporal dynamics than simple actions; and (iii) our affordance learning is based on a small training set with only weak supervision.

For the learning, we propose a Markov Chain Monte Carlo (MCMC) based algorithm to iteratively discover latent sub-events, important joints, and their functional grouping from noisy and limited training data. In particular, we design two loops in the learning algorithm, where the outer loop uses a Metropolis-Hasting algorithm to propose temporal parsing of sub-events for each interaction instance (i.e., sub-event learning), and the inner loop selects and groups joints within each type of sub-event through a modified Chinese Restaurant Process (CRP). Based on the discovered latent sub-events and affordable joints, we learn both spatial and motion potentials for grouped affordable joints in each sub-event. For the motion synthesis, we apply the learned social affordance to unseen scenarios, where one agent is assumed to be an observed human, and the other agent is assumed to be the robot that we control to interact with the observed agent (an object will be treated as part of the observation if it is also involved). To evaluate our approach, we collected a new RGB-D video dataset including 3 human-human interactions and 2 human-object-human interactions. Note that there are no human-object-human interactions in the existing RGB-D video datasets.



To our knowledge, this is the first work to study robot learning of affordances for social activities. Our work differs from the previous robot affordance learning works in the aspect that it (1) considers activities of multiple agents, (2) decomposes activities into multiple sub-events/sub-goals and learns their affordances (i.e., hierarchical affordance) that are grounded to the skeleton sequences, and (3) learns both spatial and motion affordances of multiple body-parts involved in interactions.

\subsection{Related works}


Although there are previous studies on vision-based hierarchical activity recognition \cite{Gupta2009,RyooIJCV2011,LanPAMI2012,Amer2012,Pei2013,Choi2014,Shu2015} and human-human interaction recognition \cite{Ryoo2011,Lan2014,Huang2014}, research on affordances of high-level activities has been very limited. For the robotic motion planning and object manipulation, \cite{Lee2013,Yang2015,Wu2015} presented symbolic representation learning methods for single agent activities, but low-level joint trajectories were not explicitly modeled in those works. In computer graphics, some motion synthesis approaches have been proposed \cite{Li2002,Graham2006,Wang2008,Fragkiadaki2015}, but they only learn single agent motion based on highly accurate skeleton inputs from motion capture systems.



In contrast, in this paper, we are studying affordances of dynamic agents with multiple body parts, including human-human interactions (e.g., shaking hands) as well as human-object-human interactions (e.g., object throw-catch). Its importance was also pointed out in \cite{Gibson} as ``the richest and most elaborate affordances'', and we are exploring such concept for the first time for robots. We specifically denote such affordances as \emph{social affordances}, and present an approach to learn them from human activity videos.

\section{Representation and Formulation}

\begin{figure}[t]
\centering
\begin{subfigure}{.4\textwidth}
\centering
\includegraphics[width=1.0\textwidth]{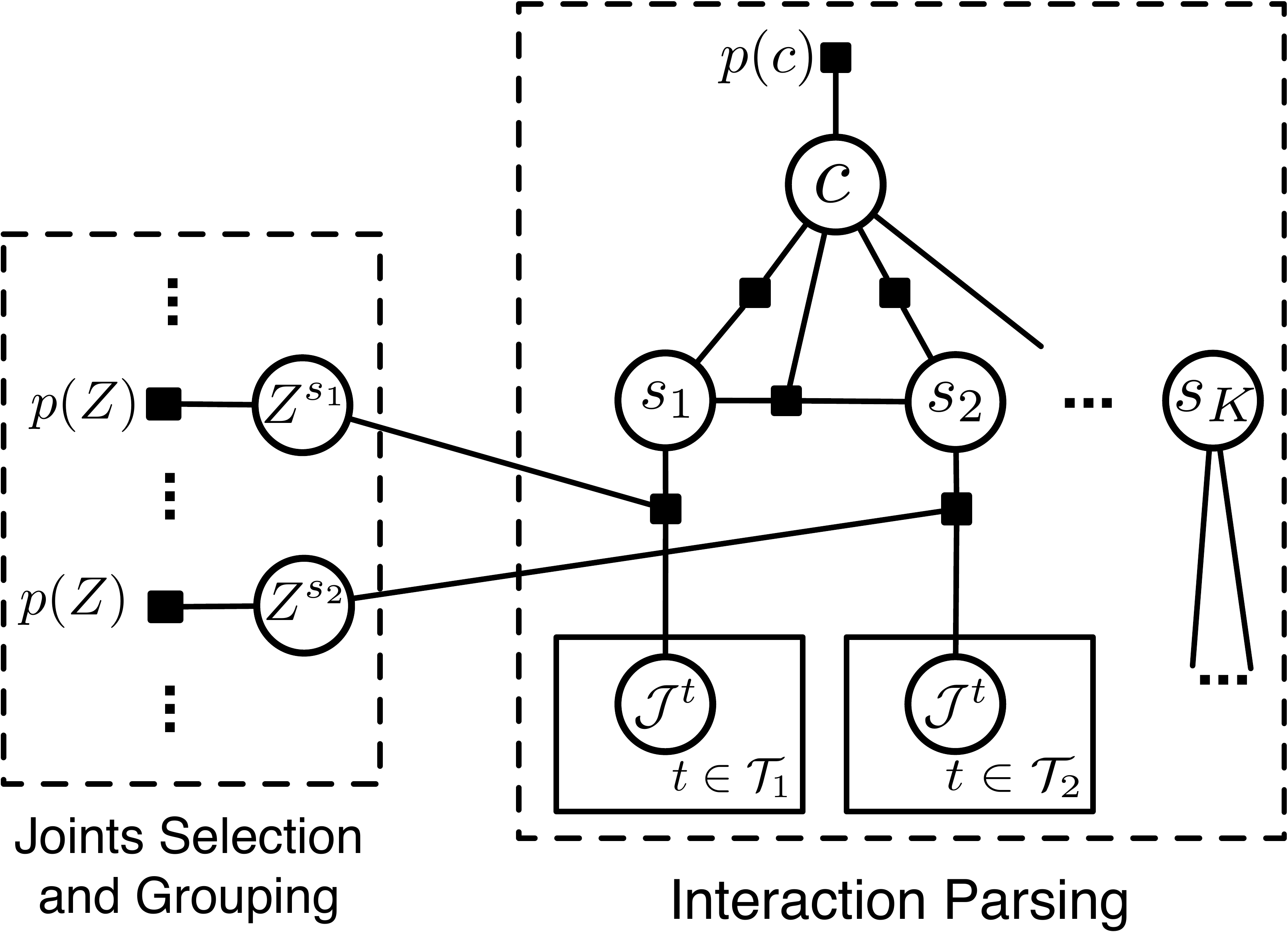}
\caption{} \label{fig:factor}
\end{subfigure}
~
\begin{subfigure}{.35\textwidth}
\centering
\includegraphics[width=1.0\textwidth]{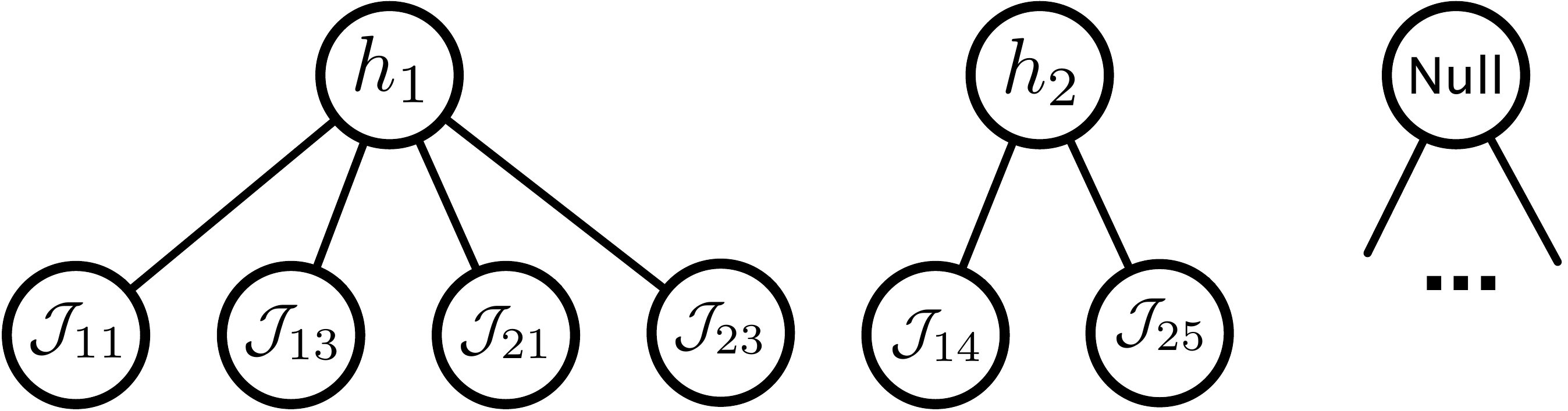}
\caption{} \label{fig:selection}
\end{subfigure}
\caption{Our model. (a) Factor graph of an interaction. (b) Selection and grouping of joints for a sub-event.}
\label{fig:model}
\end{figure}

We propose a graphical model to represent the social affordance in a hierarchical structure, which is grounded to skeleton sequences (Figure~\ref{fig:factor}). Our representation not only describes what human skeletons (i.e., body-joint locations) are likely to be observed when two persons are performing interactions, but also indicates how each interaction need to be decomposed in terms of sub-events/sub-goals and how agents should perform such sub-events in terms of joint motion.

\textbf{Skeleton sequences}. An interaction instance is represented by the skeleton sequences of the two agents.  We use $J^{t} = \{J_{1i}^{t}\} \cup \{J_{2i}^{t}\}$ to denote the positions of the two agents' joints at time $t = 1, \cdots, T$. If an interaction involves an object, then $J^{t} = \{J_{1i}^{t}\} \cup \{J_{2i}^{t}\} \cup O^t$, where $O^t$ is the position of the object at $t$. In practice, we select 5 most important joints -- base joint, left/right writs, and left/right ankles for the social affordance, whose indexes are denoted as a set $\mathcal{I}$. This reasonable simplification helps us eliminate the noise introduced by skeleton extraction from RGB-D videos while maintaining the overall characteristics of each interaction.
 
\textbf{Interaction label}. A label $c \in \mathcal{C}$ is given to an interaction to define its category, where $\mathcal{C}$ is a predefined dictionary.

\textbf{Latent sub-events}. One of our key intuitions is that a complex interaction usually consists of several steps. In order to enable the robots to mimic the human behavior, it is necessary to discover these underlying steps as latent sub-events. Here, a sub-event is defined as a sub-interval within a complete interaction. There are two crucial components in a sub-event: 1) the sub-goal to achieve at the end of the sub-event, and 2) the motion patterns to follow in this sub-event. Since it is difficult for humans to manually define and annotate the sub-events, we only specify the number of latent sub-events, i.e., $|\mathcal{S}|$, and our learning method automatically searches the optimal latent sub-event parsing for each training instance. Here, a latent sub-event parsing of an interaction instance whose length is $T$ is represented by non-overlapping intervals $\{\mathcal{T}_k\}_{k = 1, \cdots, K}$ such that $\sum_k |\mathcal{T}_k| = T$, where $\mathcal{T}_k = \{t : t = \tau^1_k, \cdots, \tau^2_k\}$, and  the sub-event labels of the corresponding intervals, i.e., $\{s_k\}_{k = 1, \cdots, K}$. Note that $K$, the number of sub-events, may vary in different instances.

 
\textbf{Joint selection and grouping}. Another key intuition of ours is to discover the affordable joints and their functional groups in each latent sub-event. This means that 1) some joints do not contribute much to accomplishing the latent sub-event due to the lack of clear motion and/or specific spatial relations among them, 2) and the rest joints are regarded as affordable joints and are further clustered together to form several functional groups, each of which has rigid spatial relations among the grouped joints in the sub-events. Figure~\ref{fig:selection} illustrates the selection and grouping of joints in a sub-event: we first select affordable joints with a Bernoulli distribution prior and remain the rest joints in a $Null$ group; then we assign each affordable joint to a functional group from a infinity number of latent functional classes $\mathcal{H} = \{h_1,\cdots,h_{\infty}\}$. The grouping can be addressed by a Chinese Restaurant Process (CRP), where a functional class is a table, and each affordable joint can be perceived as a customer to be seated at a table. We introduce auxiliary variables $Z^s = \{z_{ai}^{s} : z_{a i}^{s} \in \mathcal{H}, a \in \{1, 2\}, i = 1,\cdots, N_J\}$ to indicate the joint selection and grouping in a sub-event $s \in \mathcal{S}$ of interaction $c \in \mathcal{C}$. $J_{ai}$ is assigned to $h_{z_{ai}^s}$ if $z_{ai}^s > 0$; otherwise, $J_{ai}$ is assigned to the \textit{Null} group. Together $Z_c = \{Z^s\}_{s \in \mathcal{S}}$ represents the joint selection and grouping in a type of interaction, $c$.

\textbf{Sub-goals and motion patterns}. After grouping joints, the sub-goal of a sub-event is defined by the spatial relations (i.e., spatial potentials $\Psi_g$) among joints within the functional groups, and movements of affordable joints are described with the motion pattens (i.e., motion potentials $\Psi_m$). These allow us to infer ``how'' each agent should move.

\textbf{Parse graph}. As shown in Figure~\ref{fig:factor}, an interaction instance is represented by a parse graph $G = \langle c, S, \{J^t\}_{t = 1, \cdots, T} \rangle$. With the corresponding joint selection and grouping $Z_c$, we formalize the social affordance of an interaction as $\langle G, Z_c\rangle$. Note that $Z_c$ is fixed as common knowledge while $G$ depends on the observed instance.

\subsection{Probabilistic Modeling}
\label{sec:prob}

In this subsection, we provide how our approach models the joint probability of each parse graph $G$ and the joint selection and grouping $Z$, allowing us to use it for both (i) learning the structure and parameters of our representation based on observed human skeletons (Sec.~\ref{sec:learning}) and (ii) inferring/synthesizing new skeleton sequences for the robot using the learned model (Sec.~\ref{sec:synthesis}).


For each interaction $c$, our social affordance representation has two major parts: 1) optimal body-joint selection and grouping $Z_c$, and 2) parse graph $G$ for each observed interaction instance of $c$. Given $Z_c$, the probability of $G$ for an instance is defined as
\begin{equation}
\begin{array}{lcl}
p(G | Z_c) &\propto& \displaystyle \underbrace{\prod_k p(\{J^t\}_{t \in \mathcal{T}_k} | Z^{s_k}, s_k, c)}_{\text{likelihood}} \cdot \underbrace{p(c)}_{\text{interaction prior}} \\
&& \cdot \underbrace{\prod_{k = 2}^K p(s_k | s_{k - 1}, c)}_{\text{sub-event transition}} \cdot \underbrace{\prod_{k = 1}^K p(s_k | c)}_{\text{sub-event prior}},
\end{array}
\label{eq:g}
\end{equation}
and the prior for joint selection and grouping is
\begin{equation}
p(Z_c) = \prod_{s \in \mathcal{S}} p(Z^s | c).
\end{equation}
Hence the joint probability is
\begin{equation}
p(G, Z_c)  = p(G | Z_c) p(Z_c).
\label{eq:joint}
\end{equation}

\textbf{Likelihood}. The likelihood term in (\ref{eq:g}) consists of i) spatial potential $\Psi_g(\{J^t\}_{t \in \mathcal{T}}, Z^s, s)$ for the sub-goal in sub-event $s$, and ii) motion potential  $\Psi_m(\{J^t\}_{t \in \mathcal{T}}, Z^s, s)$ for motion patterns of the affordable joints in $s$:
\begin{equation}
\begin{array}{lcl}
 \displaystyle p(\{J^t\}_{t \in \mathcal{T}} | Z^{s}, s, c) \\
 =  \displaystyle \Psi_g(\{J^t\}_{t \in \mathcal{T}}, Z^s, s)\Psi_m(\{J^t\}_{t \in \mathcal{T}}, Z^s, s).
\end{array}
\label{eq:likelihood}
\end{equation}

\textbf{Spatial potential}. We shift the affordable joints at the end of each sub-event (i.e., $\tau^2$) in an interaction w.r.t. the mass center of the assigned functional group. The shifted joint locations at $t$ are denoted as $\widetilde{J}^{t}_{ai}$. If there is only one joint in a group, the reference point will be the base joint location of the other agent at the moment instead. Then for each joint, we have 
\begin{equation}
\psi_g(\widetilde{J}^t_{ai}) = \psi_{\text{xy}}(\widetilde{J}^t_{ai})\psi_{\text{z}}(\widetilde{J}^t_{ai})\psi_{\text{o}}(\widetilde{J}^t_{ai}),
\end{equation}
 where $\psi_{\text{xy}}(\widetilde{J}^t_{ai})$ and $\psi_{\text{z}}(\widetilde{J}^t_{ai})$ are Weibull distributions of the horizontal and vertical distance between the joint and the reference point, and $\psi_{\text{o}}(\widetilde{J}^t_{ai})$ is a von Mises distribution for the angle between the two points. Note that the spatial potential only accounts for affordable joints (i.e., $z^s_{ai} > 0$). Thus 
 \begin{equation}
 \Psi_{g}(\{J^t\}_{t \in \mathcal{T}}, Z^s, s) = \prod_{a,i}\psi_g(\widetilde{J}^{\tau^2}_{ai})^{\mathds{1}(z^s_{ai} > 0)}.
 \end{equation}

\textbf{Motion potential}. In a sub-event $s$ of an interaction, we compute the movement of a joint $J_{ai}$ by $\db_{ai} = J^{\tau^2}_{ai} - J^{\tau^1}_{ai}$. Similar to the spatial potential, this joint's motion potential is 
\begin{equation}
\psi_m(\{J_{ai}^t\}_{t \in \mathcal{T}}) = \psi_m(\db_{ai}) = \psi_{\text{xy}}(\db_{ai})\psi_{\text{z}}(\db_{ai})\psi_{\text{o}}(\db_{ai}).
\label{eq:motionP}
\end{equation}
For an affordable joint, we use Weibull distributions for both horizontal and vertical distances and a von Mises distribution for the orientation. To encourage static joints to be assigned to the $Null$ group, we fit exponential distributions for the distances while keeping $\psi_{\text{o}}(\db_{ai})$ the same if $z_{ai}^s = 0$. Hence,
\begin{equation}
\Psi_m(\{J^t\}_{t \in \mathcal{T}}, Z^s, s) = \prod_{a,i} \psi_m(\{J_{ai}^t\}_{t \in \mathcal{T}_k}).
\end{equation}

\textbf{Prior for interaction category and sub-event transition}. We assume uniform distribution for $p(c)$ and compute the transition frequency from training data for $p(s_k | s_{k - 1}, c)$.

\textbf{Sub-event prior}. The duration of a sub-event $s_k$ in interaction $c$ is regularized by a log-normal distribution $p(s_k | c)$:
\begin{equation}
\displaystyle p(s_k | c) = \exp\{-(\ln |\mathcal{T}_k|-\mu)^2/(2\sigma^2)\} / (|\mathcal{T}_k|\sigma\sqrt{2\pi}).
\label{eq:durationPrior}
\end{equation}

\textbf{Joint selection and grouping prior}. Combined with Bernoulli distribution and the prior of CRP, the joint selection and grouping prior for $Z^s$ in sub-event type $s$ of interaction $c$ is defined as
\begin{equation}
\displaystyle p(Z^s | c) = \underbrace{\dfrac{\prod_{h} (M_h -1)!}{M!}}_{\text{CRP prior}} \prod_{ai} \underbrace{\beta^{\mathds{1}(z^s_{ai} > 0)} (1 - \beta)^{(1 - \mathds{1}(z^s_{ai} > 0))}}_{\text{Bernoulli prior for a joint}}.
\label{eq:jointprior}
\end{equation}
where $M_h$ is the number of joints assigned to latent function group $h$, and $M$ is the total number of affordable joints, i.e., $M = \sum_{a,i} \mathds{1}(z^s_{ai} > 0)$.


\begin{algorithm}[t]
\caption{Learning Algorithm}
\label{alg:learning}
\begin{algorithmic}[1]
\small
\State Input: $\{J^t\}_{t = 1, \cdots, T}$ of each instance with the same interaction category $c \in \mathcal{C}$
\State Obtain the atomic time intervals by K-means clustering
\State Initialize $S$ of each instance, and $Z_c$
\Repeat
\State Propose $S^\prime$
\Repeat
\State Sample new $Z_c$ through Gibbs sampling
\Until{Convergence}
\State $\alpha = \min\{\frac{Q(S^\prime \rightarrow S)P^*(\mathcal{G}^\prime, Z_c^\prime)}{Q(S \rightarrow S^\prime)P^*(\mathcal{G}, Z_c)}, 1\}$
\State $u \sim Unif[0, 1]$
\State If $u \leq \alpha$, accept the proposal $S^\prime$
\Until{Convergence}
\end{algorithmic}
\end{algorithm}

\section{Learning}
\label{sec:learning}

Given the skeleton sequences and their interaction labels, we learn the model for each interaction category in isolation. Assume that we have $N$ training instances for interaction $c$, then will have $N$ parse graphs $\mathcal{G} = \{G^n\}_{n = 1, \dots, N}$, and a common $Z_c$ for this type of interaction. The objective of our learning algorithm is to find the optimal $\mathcal{G}$ and $Z_c$ that maximize the following joint probability: 
\begin{equation}
p(\mathcal{G}, Z_c) = p(Z_c)\prod_n^N p(G^n | Z_c).
\end{equation}
Note that the size of latent sub-event dictionary, $|\mathcal{S}|$, is specified for each interaction.

We propose a MCMC learning algorithm as Alg.~\ref{alg:learning}, which includes two optimization loops:
\begin{itemize} 
\itemsep-0em 
\item[1] Metropolis-Hasting algorithm for sub-event parsing. 
\item[2] Given sub-event parsing, apply Gibbs sampling for the optimization $Z^*_c = \argmax_{Z_c} p(\mathcal{G}, Z_c) = \argmax_{Z_c} p(\mathcal{G} | Z_c) p(Z_c)$.
\end{itemize}

The details of two loops are introduced as follows.

\subsection{Outer Loop for Sub-Event Parsing}
In the outer loop, we optimize the sub-event parsing by a Metropolis-Hasting algorithm. We first parse each interaction sequence into atomic time intervals using K-means clustering of agents' skeletons (we use 50 clusters). Then the sub-events are formed by merging some of the atomic time intervals together. At each iteration, we propose a new sub-event parsing $S^\prime$ through one of the following dynamics: 


\textbf{Merging}. In this dynamics, we merge two sub-events with similar skeletons together and uniformly sample a new sub-event label for it, which forms a new sub-event parsing $S^\prime$. For this, we first define the distance between two consecutive sub-events by the mean joint distance between the average skeletons in these two sub-events, which is denoted by $d$. Then the proposal distribution is $Q(S \rightarrow S^\prime | d) \propto e^{-\lambda d} / N_L$, where $\lambda$ is a constant number, and $N_L$ is number of possible label assignments for the new sub-event. In practice, we set $\lambda = 1$.

\textbf{Splitting}. We can also split a sub-event with multiple atomic time intervals into two non-overlapping sub-events with two new labels. Note that an atomic time interval is not splittable. Similarly, we can compute the distance $d$ between the average skeletons of these two new sub-events and assume uniform distributions for the new labels. To encourage the split of two sub-events with very different skeletons, we define the proposal distribution to be $Q(S \rightarrow S^\prime | d) \propto (1 - e^{-\lambda d}) / N_L$, where $N_L$ is number of possible new labels.

\textbf{Re-labeling}. We relabel a uniformly sampled sub-event for this dynamics, which gives the proposal distribution $Q(S \rightarrow S^\prime | d) = 1 / (N_L \cdot N_S)$, where $N_L$ and $N_S$ are the numbers of possible labels and current sub-events respectively.

In addition, the type of dynamics at each iteration is sampled w.r.t. these three probabilities, $q_1 = 0.4$, $q_2 = 0.4$, $q_3 = 0.2$, for the above three types respectively.

The acceptance rate $\alpha$ is then defined as $\alpha = \min\{\frac{Q(S^\prime \rightarrow S)P^*(\mathcal{G}^\prime, Z_c^\prime)}{Q(S \rightarrow S^\prime)P^*(\mathcal{G}, Z_c)}, 1\}$, where $P^*(\mathcal{G}, Z_c)$ is the highest joint probability given current sub-event parsing $S$, i.e., $P^*(\mathcal{G}, Z_c) = \max_{Z_c} p(\mathcal{G}, Z_c)$ . Similarly, $P^*(\mathcal{G}^\prime, Z_c^\prime) = \max_{Z_c^\prime} p(\mathcal{G}^\prime, Z_c^\prime)$. 

\subsection{Inner Loop for Joint Selection and Grouping}

To obtain $P^*(\mathcal{G}^\prime, Z_c^\prime)$ in the acceptance rate defined for the outer loop given the proposed $S^\prime$, we use Gibbs sampling to iteratively update $Z_c^\prime$. At each iteration, we assign a joint from $\mathcal{I}$ to a new group in each type of sub-event by
\begin{equation}
z_{ai}^s \sim p(\mathcal{G} | Z_c^\prime) p(z_{ai}^s | Z^s_{-ai}).
\end{equation}
Based on (\ref{eq:jointprior}), we have
\begin{equation}
p(z_{ai}^s | Z^s_{-ai}) = 
\begin{cases}
\beta\dfrac{\gamma}{M - 1 + \gamma} & \text{if } z_{ai}^s > 0, M_{z_{ai}^s} = 0 \\
\beta\dfrac{M_{z_{ai}^s}}{M - 1 + \gamma} & \text{if } z_{ai}^s > 0, M_{z_{ai}^s} > 0 \\
1 - \beta & \text{if } z_{ai}^s = 0
\end{cases}
\end{equation}
where the variables have the same meaning as in (\ref{eq:jointprior}) and $\beta = 0.3$ and $\gamma = 1.0$ are the parameters for our CRP.

\begin{algorithm}[t]
\caption{Motion Synthesis Algorithm}
\label{alg:simulation}
\begin{algorithmic}[1]
\small
\State Give the interaction label $c$ and the total length $T$; set unit time interval for simulation to be $\Delta T = 5$; input the skeletons in the first $T_0 = 10$ frames, i.e., $\{J^t\}_{t = 1, \cdots, T_0}$; set $\tau \leftarrow T_0$
\Repeat
\State Input $\{J_{1i}\}^t_{t = \tau + 1, \cdots, \tau + \Delta T}$
\State Extend $\{J_{2i}\}^t$ to $\tau + \Delta T$ by copying $\{J_{2i}\}^\tau$ temporarily
\State Infer $S$ of $\{J^t\}_{t = 1, \cdots, \tau + \Delta T}$ by DP; we assume that the last sub-event, $s_K$, is the current on-going sub-event type
\State Predict the ending time $\tau_K^2$ of $s_K$ by sampling the complete duration $|\mathcal{T}|$ w.r.t. the prior defined in (\ref{eq:durationPrior}), and generate $N = 100$ possible samples for the locations of the modeled five joints in $\mathcal{I}$, i.e., $\{\hat{J}_{2i^\prime}^n\}_{i^\prime \in \mathcal{I}, n = 1, \cdots, N}$; note that the joints in the \textit{Null} group are assumed to be static in the current sub-event
\State Obtain the $N$ corresponding joint locations at current time $\tau + \Delta T$, $\{J_{2i^\prime}^n\}_{i^\prime \in \mathcal{I}, n = 1, \cdots, N}$, by interpolation based on $\{\hat{J}_{2i^\prime}^n\}$
\State We choose the one that maximizes the likelihood, i.e., $\{J_{2i^\prime}^*\}_{i^\prime \in \mathcal{I}}$, by computing motion and spatial potentials
\State Fit clustered full body skeletons from K-means to $\{J_{2i^\prime}^*\}_{i^\prime \in \mathcal{I}}$ by rotating limbs, and obtain the closest one $\{J_{2i}^*\}$
\State $J_{2i}^{\tau + \Delta T} \leftarrow J_{1i}^*$
\State Interpolate the skeletons from $\tau + 1$ to $\tau + \Delta T $, and update $\{J_{2i}\}^t_{t = \tau + 1, \cdots, \tau + \Delta T}$
\State $\tau \leftarrow \tau + \Delta T$
\Until{$\tau \geq T$}
\end{algorithmic}
\end{algorithm}

\section{Motion Synthesis}
\label{sec:synthesis}

Our purpose for learning social affordance is to teach a robot how to interact with a human. Hence, we design an online simulation method to ``synthesize'' a skeleton sequence (i.e., $\{J_{2i}\}^t_{t = 1, \cdots, T}$) as a robot's action sequence to interact with a human (i.e., the first agent) and an object given the observed skeleton sequence (i.e., $\{J_{1i}\}^t_{t = 1, \cdots, T}$), where $T$ is the length of the interaction. The idea is to make our approach automatically ``generate'' an agent's body joint motion based on the learned social affordance and the other agents' motion. Note that the human skeleton sequence has not been seen in the training data and we assume that the interaction category $c$ is given. The estimated object trajectory $\{O^t\}_{t = 1, \cdots, T}$ will also be used if an object is involved. Since we define the social affordance for a interaction instance as $\langle G, Z_c\rangle$, the synthesis is essentially to infer the joint locations for the second agent (i.e., $\{J_{2i}\}^t$) by maximizing the joint probability defined in (\ref{eq:joint}).

The main steps of our motion synthesis are summarized in Alg.~\ref{alg:simulation}. At any time $t$, we first use a dynamic programming (DP) algorithm to estimate current sub-event type based on our observations of the human agent (and the object if it exists) as well as the skeletons that we have synthesized so far. Then we sample the new joint locations by maximizing the spatial and motion potentials under current sub-event.



\subsection{Dynamic Programming}
We use the following DP algorithm to efficiently infer the latent sub-events given the skeletons of two agents (and the object trajectory if present) by maximizing the probability of the parse graph defined in (\ref{eq:g}). For a sequence of interaction $c$, we first define $m(s^\prime, t^\prime, s, t)$ as the log probability of assigning sub-event type $s$ to the time interval $[t^\prime, t]$ when the preceding sub-event type is $s^\prime$, which can be computed as
\begin{equation}
\begin{array}{lcl}
m(s^\prime, t^\prime, s, t) &=& \log p(\{J^t\}_{t \in [t^\prime, t]} | Z^{s}, s, c) \\
&& + \log p(t - t^\prime + 1 | s, c) + \log p(s | s^\prime, c)
\end{array}
\end{equation}
Then we define the highest log posterior probability for assigning type $s$ to the last sub-event of $\{J^t\}_{t = 1, \cdots, t}$ as $b(s, t)$:
\begin{equation}
b(s, \tau) = \max_{s^\prime \neq s, t^\prime < t} \{b(s^\prime, t^\prime) +  m(s^\prime, t^\prime, s, t)\}
\label{eq:belief}
\end{equation}
where $b(s, 0) = 0$. By recording all pairs of $s^\prime$ and $t^\prime$ that maximize $b(s, t)$ in (\ref{eq:belief}), we can easily backtrace the optimal latent sub-event parsing including labels $s_1, \cdots, s_K$ and corresponding intervals $\mathcal{\mathcal{T}}_1, \cdots, \mathcal{\mathcal{T}}_K$, starting from the last frame until the first frame in a reverse process.

\begin{figure}[t]
\centering
\includegraphics[width=0.48\textwidth]{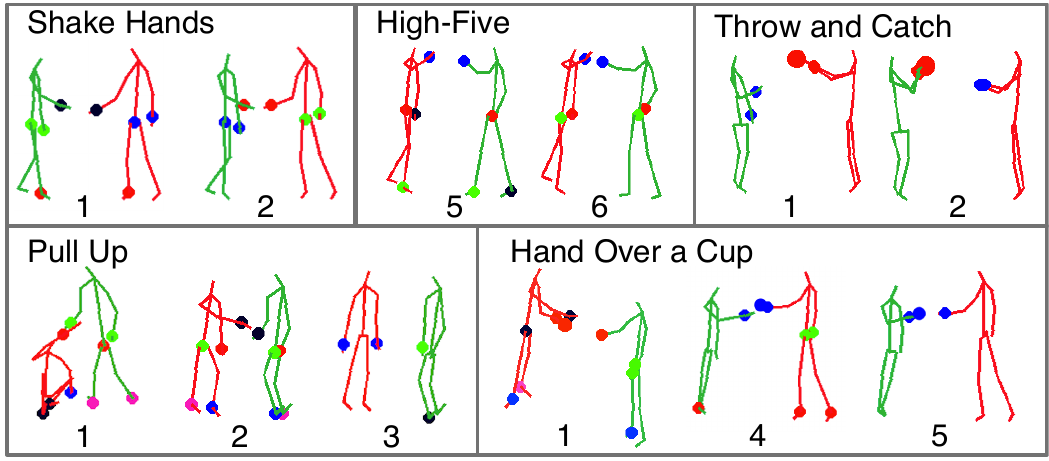}
\caption{Visualization of some discovered sub-events and their joint grouping in the five interactions, where the number denotes the sub-event label and the joint colors show the groups. For \textit{throw and catch} and \textit{hand  over a cup}, an object is also displayed as an additional affordable joint. The shown frames are the last moments of the corresponding sub-events, which depict the learned sub-goals.}
\label{fig:subevent}
\end{figure}

\begin{figure*}[t]
\centering
\includegraphics[width=1.0\textwidth]{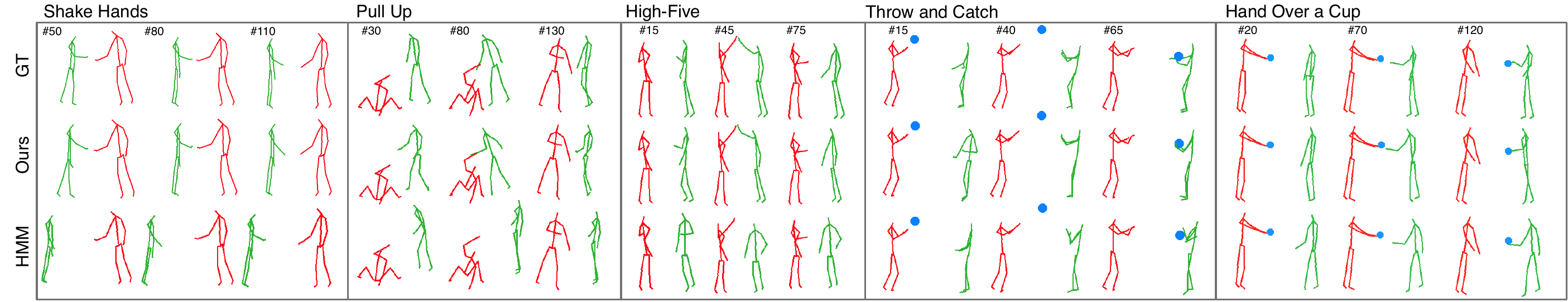}
\caption{Comparison between synthesized and GT skeletons. The red agent and the blue object are observed; the green agents are either GT skeletons, synthesized skeletons by ours, or those by HMM respectively. The numbers are the frame indexes.}
\label{fig:qualitative}
\end{figure*}


\section{Experiment}


We collected a new RGB-D video dataset, i.e., UCLA Human-Human-Object Interaction (HHOI) dataset, which includes 3 types of human-human interactions, i.e., \textit{shake hands}, \textit{high-five}, \textit{pull up}, and 2 types of human-object-human interactions, i.e., \textit{throw and catch}, and \textit{hand over a cup}. On average, there are 23.6 instances per interaction performed by totally 8 actors recorded from various views. Each interaction lasts 2-7 seconds presented at 10-15 fps. We used the MS Kinect v2 sensor for the collection, and also took advantage of its skeleton estimation. The objects are detected by background subtraction on both RGB and depth images. The dataset is available at: \url{http://www.stat.ucla.edu/~tianmin.shu/SocialAffordance}.


We split the instances by four folds for the training and testing where the actor combinations in the testing set are different from the ones in the training set. For each interaction, our training algorithm converges within 100 outer loop iterations, which takes 3-5 hours to run on a PC with an 8-core 3.6 GHz CPU. Our motion synthesis can be ran at the average speed of 5 fps with our unoptimized Matlab code.



{\flushleft\textbf{Experiment 1:} Our approach learns affordance representations from the training set, and uses the testing set to ``synthesize'' the agent (i.e., robot) skeletons in reaction to the interacting human skeletons (and an object). We first measured the average joint distance between synthesized skeletons and the ground truth (GT) skeletons since good synthesis should not be very different from GT. A multi-level hidden Markov model (HMM) is implemented as the baseline method, where the four levels from top to bottom are: 1) the quantized distance between agents, 2) the quantized relative orientation between agents, 3) the clustered status of the human skeleton and the object, and 4) the clustered status of the synthesized skeleton. In addition, we also compare our full model with a few variants: ours without joint selection and grouping (V1), and ours without the latent sub-events (V2). Notice that this social affordance based skeleton synthesis is a new problem and we are unaware of any exact prior state-of-the-art approach.}

The average joint distance for different methods are compared in Table.~\ref{table:quantitative}. Our full model outperforms all other approaches by a large margin, which proves the advantage of our hierarchical generative model with latent sub-events and joint grouping. Note that the tracking error of Kinect 2 for a joint ranges from 50 mm and 100 mm \cite{Wang2015}. Figure~\ref{fig:subevent} demonstrates a few joint selection and grouping results for some automatically discovered latent sub-events in different interactions. We also visualize several synthesized interactions in Figure~\ref{fig:qualitative}, where the synthesized skeletons from ours and the HMM baseline are compared with GT skeletons.


\begin{table}[t!]\scriptsize
\centering
\tabcolsep=0.06cm
\begin{tabular}{c||c|c|c|c|c||c} \hline
Method   & Shake Hands & Pull Up & High-Five & Throw \& Catch & Hand Over & Average \\ \hline
HMM &   $0.362$ & $ 0.344$ & $0.284$ & $0.189$ & $0.229$ & $0.2816$     \\ \hline
V1  &   $0.061$ & $ 0.144$ & $0.079$ & $0.091$ & $0.074$ & $0.0899$    \\ \hline
V2 & $0.066$ & $ 0.231$ & $0.090$ & $0.109$ & $0.070$ & $0.1132$\\ \hline
Ours & $\bm{0.054}$ & $\bm{0.109}$ & $\bm{0.058}$ & $\bm{0.076}$ & $\bm{0.068}$ & $\bm{0.0730}$ \\ \hline
\end{tabular}
\caption{Average joint distance (in meters) between synthesized skeletons and GT skeletons for each interaction.}
\label{table:quantitative}
\end{table}

{\flushleft\textbf{Experiment 2:} In addition, we also conducted a user study experiment of comparing the naturalness of our synthesized skeleton vs. ground truths. Similar to \cite{Meisner2009}, we asked 14 human subjects (undergraduate/graduate students at UCLA) to rate the synthesized and GT interactions. For this, we predefined 4 sets of videos, where there were 5 videos for each interaction in a set, and all these 5 videos were either from GT or ours. Thus each set had a mixture of videos of GT and ours, but GT and ours did not co-exist for any interaction. Then we randomly assigned these 4 sets to the subjects who were asked to watch each video in the given set only once and rate it from 1 (worst) to 5 for three different questions: ``Is the purpose of the interaction \emph{successfully} achieved?'' (Q1), ``Is the synthesized agent behaving \emph{naturally}?'' (Q2), and ``Does the synthesized agent look like a \emph{human} rather than a robot?'' (Q3). The subjects were instructed that the red skeleton was a real human and the green skeleton was synthesized in all videos. They were not aware of the fact that GT and our synthesized sequences were mixed in the stimuli.}

\begin{table}[t!]\scriptsize
\centering
\tabcolsep=0.04cm
\begin{tabular}{c|c|c|c|c|c|c} \hline
 & Source   & Shake Hands & Pull Up & High-Five & Throw \& Catch & Hand Over \\ \hline
\multirow{2}{*}{Q1} & Ours &   $\bm{4.60} \pm 0.69$ & $ 3.90 \pm 0.70 $ & $4.53 \pm 0.30$ & $\bm{4.31} \pm 0.89$ & $4.40 \pm 0.37$     \\ \cline{2-7}
 & GT &   $4.50 \pm 0.82$ & $ 4.29 \pm 0.58 $ & $4.64 \pm 0.33$ & $4.20 \pm 0.76$ & $4.64 \pm 0.30$    \\ \hline
\multirow{2}{*}{Q2}  & Ours & $\bm{4.23} \pm 0.34$ & $ 2.80 \pm 0.75$ & $3.70 \pm 0.47$ & $\bm{4.06} \pm 0.83$ & $3.89 \pm 0.38$ \\ \cline{2-7}
 & GT & $4.20 \pm 0.47$ & $4.23 \pm 0.48$ & $4.64 \pm 0.17$ & $3.86 \pm 0.53$ & $4.24 \pm 0.46$ \\ \hline
\multirow{2}{*}{Q3}  & Ours & $4.23 \pm 0.50$ & $ 2.63 \pm 0.60$ & $3.57 \pm 0.73$ & $\bm{4.03} \pm 0.88$ & $3.69 \pm 0.64$ \\ \cline{2-7}
 & GT & $4.30 \pm 0.60$ & $3.71 \pm 1.15$ & $4.40 \pm 0.63$ & $3.97 \pm 0.74$ & $4.40 \pm 0.24$ \\ \hline
\end{tabular}
\caption{The means and standard deviations of human ratings for the three questions. The highlighted ratings indicate that the sequences synthesized by ours have higher mean ratings than GT sequences.}
\label{table:ratings}
\end{table}


Table \ref{table:ratings} compares the mean and standard deviation of human ratings per interaction per question. Following \cite{Walker2011}, we test the equivalence between the ratings of ours and GT for each question using $90\%$ confidence interval. When the equivalence margin is $0.5$, \textit{shake hands} and \textit{throw and catch} pass the test for all three questions while the rest interactions only pass the test for Q1. When we consider the equivalence margin to be $1$, only \textit{pull up} does not pass the equivalence test for Q2 and Q3. Overall, our motion synthesis is comparable to Kinect-based skeleton estimation, especially for Q1, suggesting that we are able to learn an appropriate social affordance representation. The lower ratings for \textit{pull up} mainly results from much noisier training sequences. Interestingly, the synthesized sequences of \textit{shake hands} and \textit{throw and catch} have sightly higher ratings than GT for Q1 and Q2. This is because our model learns affordances from multiple training sequences, whereas GT is based on a single and noisy Kinect measure. One distinguishable effect is hand touching, which is a critical pattern for the human subjects to rate the videos according to their feedback after the experiment. In GT videos, especially \textit{shake hands} and \textit{throw and catch}, the hand touching (either with another agent's hand or the ball) is not captured due to occlusion, whereas our synthesized skeletons have notably better performances since our method automatically groups the corresponding wrist joints (and the ball) together to learn their spatial relations, as shown in Figure~\ref{fig:qualitative}. This shows that our approach is learning sub-goals of the interactions correctly even with noisy Kinect skeletons. 

For Q3, we also counted the frequencies of the high scores (4 or 5) given to the five interactions: 0.87, 0.17, 0.53, 0.77, 0.63 for ours, and  0.88, 0.69, 0.84, 0.66, 0.84 for GT respectively (ordered as in Table \ref{table:ratings}). This is similar to the Turing test: we are measuring whether the subjects perceived the agent as more human-like or more robot-like.

After synthesizing the skeleton sequence, applying the social affordances learned from human activities to the robot replication is straightforward. Since we explicitly represent the spatial and motion patterns of the base joint and the end points of the limbs, we can match them to the corresponding base position and end positions of limbs on a robot. Consequently movement control of these key positions of a robot can be achieved by moving them based on the synthesized trajectories of their human joint counterparts to reach the desired sub-goals. We will implement this on a real robotic system in the future work.

\section{Conclusion}

In this paper, we discussed the new concept of \emph{social affordance}. We were able to confirm that our approach learns affordance on human body-parts from human interactions, finding important body joints involved in the interactions, discovering latent sub-events, and learning their spatial and motion patterns. We also confirmed that we are able to synthesize future skeletons of agents by taking advantage of the learned affordance representation, and that it obtains results comparable to RGBD-based ground truth skeletons estimated from Kinect.

One future work is to transfer our learned human motion model to a robot motion model. In this paper, we focused on the affordance ``learning'' part, and we took advantage of it to synthesize skeleton motion sequences by assuming that humans and robots share their body configurations and motion (i.e., a humanoid robot). However, in practice, robots have different configurations and mechanical constraints than humans. In order for the learned social affordance to be useful for robots in general (e.g., non-humanoid robots), motion transfer is needed as a future research challenge.

\section*{Acknowledgments}

This research has been sponsored in part by grants DARPA SIMPLEX project N66001-15-C-4035 and ONR MURI project N00014-16-1-2007. 

\bibliographystyle{named}
\bibliography{ijcai16}

\end{document}